\pdfoutput=1

\documentclass[11pt]{article}

\usepackage[final]{acl}

\usepackage{times}      
\usepackage{latexsym} 

\usepackage[T1]{fontenc}

\usepackage[utf8]{inputenc}

\usepackage{microtype}

\usepackage{inconsolata}

\usepackage{graphicx}

\usepackage{amsmath} 
\usepackage{tcolorbox}
\usepackage{enumitem}
\usepackage{array}
\usepackage{tabularx}
\usepackage{booktabs}
\usepackage{multirow}
\usepackage{subfigure}
\usepackage{bm} 

\def\chkp#1{{\texttt{#1}}} 

\newcommand{\ip}[1]{\langle #1 \rangle}

\def\topic#1{{``\textcolor{OliveGreen}{\textit{#1}}''}}   
\def\leftind#1{{``\textcolor{NavyBlue}{\textit{#1}}''}}   
\def\rightind#1{{``\textcolor{BrickRed}{\textit{#1}}''}}  

\renewcommand{\paragraph}[1]{\smallskip\noindent\textbf{#1.}}
\renewcommand{\subparagraph}[1]{\smallskip\noindent\textbf{\underline{#1.}}}

\newcolumntype{L}[1]{>{\raggedright\arraybackslash}p{#1}}  

\title{PRISM: A Framework for Producing Interpretable Political Bias Embeddings with Political-Aware Cross-Encoder}

\author{
 \textbf{Yiqun Sun\textsuperscript{1}},
 \textbf{Qiang Huang\textsuperscript{2}\thanks{Qiang Huang is the corresponding author.}},
 \textbf{Anthony K. H. Tung\textsuperscript{1}},
 \textbf{Jun Yu\textsuperscript{2}}
\\
 \textsuperscript{1}School of Computing, National University of Singapore
\\
 \textsuperscript{2}School of Intelligence Science and Engineering, Harbin Institute of Technology (Shenzhen)
\\
   \{sunyq, atung\}@comp.nus.edu.sg,~\{huangqiang, yujun\}@hit.edu.cn
}

\begin{document}

\maketitle
\begin{abstract}
Semantic Text Embedding is a fundamental NLP task that encodes textual content into vector representations, where proximity in the embedding space reflects semantic similarity. 
While existing embedding models excel at capturing general meaning, they often overlook ideological nuances, limiting their effectiveness in tasks that require an understanding of political bias. 
To address this gap, we introduce \textbf{PRISM}, the first framework designed to \textbf{P}roduce inte\textbf{R}pretable pol\textbf{I}tical bia\textbf{S} e\textbf{M}beddings. 
PRISM operates in two key stages: 
(1) Controversial Topic Bias Indicator Mining, which systematically extracts fine-grained political topics and their corresponding bias indicators from weakly labeled news data, and 
(2) Cross-Encoder Political Bias Embedding, which assigns structured bias scores to news articles based on their alignment with these indicators. 
This approach ensures that embeddings are explicitly tied to bias-revealing dimensions, enhancing both interpretability and predictive power. 
Through extensive experiments on two large-scale datasets, we demonstrate that PRISM outperforms state-of-the-art text embedding models in political bias classification while offering highly interpretable representations that facilitate diversified retrieval and ideological analysis. 
The source code is available at \url{https://github.com/dukesun99/ACL-PRISM}.
\end{abstract}

\section{Introduction}
\label{sec:intro}

Semantic Text Embedding is a fundamental NLP task that encodes texts into vector representations, where proximity in the embedding space reflects semantic similarity.
This task has garnered extensive research attention \citep{mikolov2013distributed, pennington-etal-2014-glove, devlin-etal-2019-bert, gao-etal-2021-simcse, zhuo-etal-2023-whitenedcse, li-li-2024-aoe} due to its broad applications in retrieval \citep{karpukhin-etal-2020-dense, thakur2021beir}, clustering \citep{aggarwal2012survey}, and semantic textual similarity (STS) \citep{agirre-etal-2012-semeval, agirre-etal-2013-sem}. 

\begin{figure}[t]
  \centering
  \captionsetup{skip=0.75em,belowskip=-0.75em}
  \includegraphics[width=0.99\columnwidth]{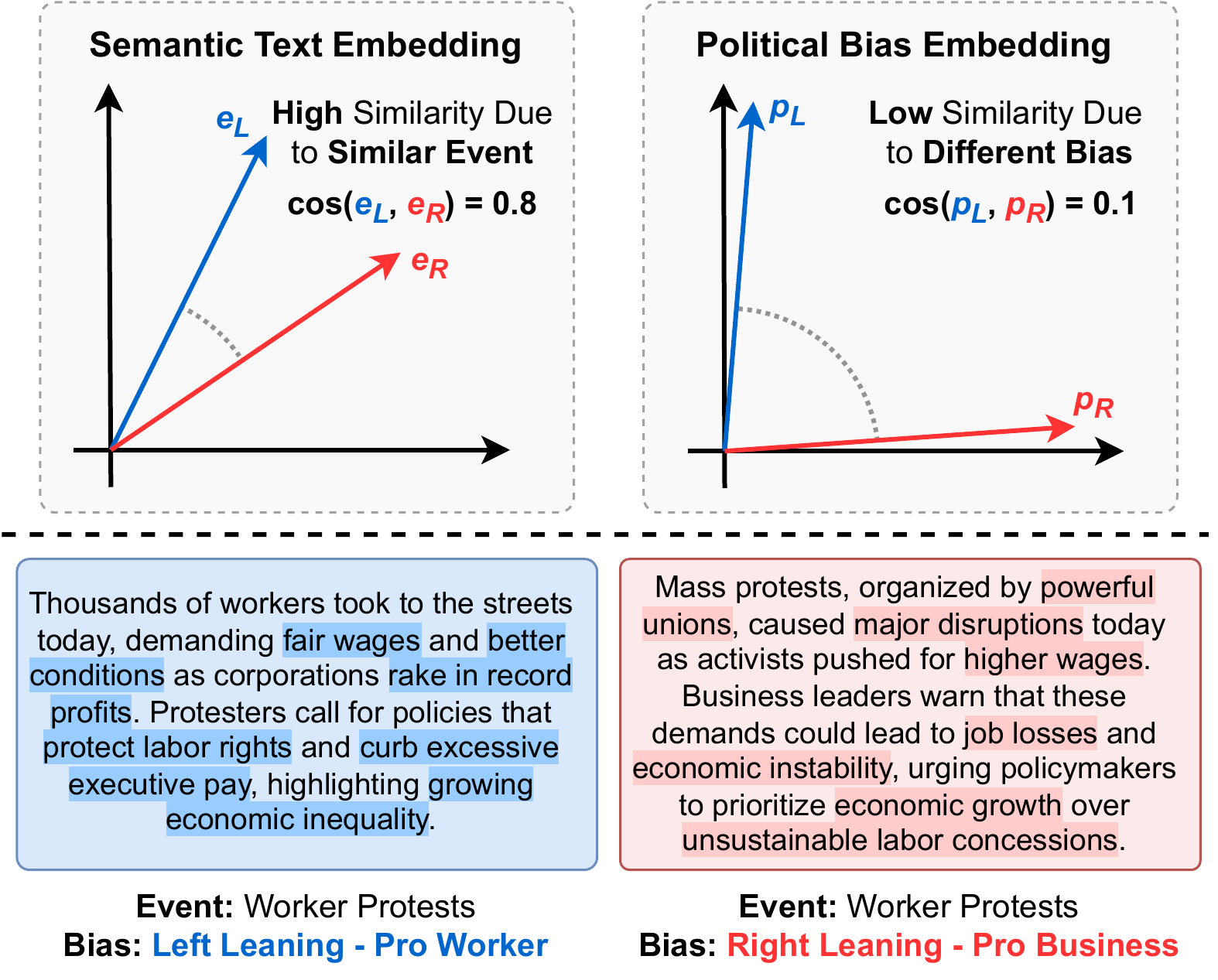}
  \caption{Semantic Text Embedding vs. Political Bias Embedding: While the former captures event-level similarity, the latter reveals ideological orientation in news coverage.}
  \label{fig:example}
\end{figure}

While significant progress has been made in developing advanced models for semantic representation, these approaches primarily capture general meaning while often failing to account for underlying political bias. 
This limitation becomes apparent in real-world scenarios such as news coverage of the same event--illustrated in Figure \ref{fig:example}--where articles may report on identical topics but convey distinctly different political perspectives. 
Existing embedding models \citep{gao-etal-2021-simcse, zhuo-etal-2023-whitenedcse, li-li-2024-aoe}, despite assigning high similarity scores based on shared content, struggle to capture these ideological nuances, exposing a crucial gap in current methodologies.

To address this gap, this work investigates \textbf{Political Bias Embedding}, a new representation learning task that transforms textual content into compact vector representations customized to capture ideological orientations.  
These specialized embeddings offer significant advantages for various downstream applications, ranging from political bias classification \citep{iyyer-etal-2014-political, liu-etal-2022-politics} to diversified retrieval systems that expose content across the ideological spectrum \citep{sun2024diversinews, huang2024diversity}. 

Existing research on political bias analysis has evolved from simple left-right classification~\citep{iyyer-etal-2014-political, baly-etal-2020-detect} to more nuanced approaches that consider multiple finer-grained political dimensions~\citep{kim-johnson-2022-close, liu-etal-2023-ideology}. 
In parallel, advances in text embeddings have progressed from general-purpose representations~\cite{reimers-gurevych-2019-sentence} to domain-specific models tailored for specialized fields~\cite{lee2020biobert, beltagy-etal-2019-scibert}. 
Specifically, for political text analysis, recent work has developed specialized encoder-only Transformer models \cite{devlin-etal-2019-bert, liu2019roberta} through fine-tuning with political news articles~\cite{liu-etal-2022-politics}, while instruction-following text embedding models have emerged as another promising direction~\cite{su-etal-2023-one, peng-etal-2024-answer}. 

Despite these advancements, developing effective political bias embeddings remains challenging.
\begin{itemize}[nolistsep,left=2pt]
  \item \textbf{Complexity of Political Dimensions:} 
  While political bias analysis has moved beyond binary classification toward multi-dimensional perspectives~\citep{kim-johnson-2022-close, liu-etal-2023-ideology}, enumerating a comprehensive political bias taxonomy remains challenging.
  Automatically extracting fine-grained political topics from a news corpus is essential for learning robust and interpretable representations of political bias.
  
  \item \textbf{Scarcity of Fine-Grained Annotations:} 
  High-quality, large-scale datasets capturing nuanced political perspectives are scarce, as manually annotating political bias is resource-intensive and subjective \citep{sinno-etal-2022-political, kim-johnson-2022-close, liu-etal-2023-ideology}.
  
  \item \textbf{Black-box Models:} 
  While recent studies have explored interpretable embeddings using yes/no questions as dimensions \cite{mcinerney-etal-2023-chill, benara2024crafting, sun2025general}, these approaches are not directly applicable to political bias analysis due to the complexity and subjectivity of political viewpoints. 
  To the best of our knowledge, no existing work has systematically addressed the challenge of constructing interpretable embeddings for political bias.
\end{itemize} 

To overcome these challenges, we introduce \textbf{PRISM}, the first framework developed to \textbf{P}roduce inte\textbf{R}pretable pol\textbf{I}tical bia\textbf{S} e\textbf{M}bedding. 
PRISM comprises two core stages:
\begin{itemize}[nolistsep,left=2pt]
  \item \textbf{Controversial Topic Bias Indicator Mining:} 
  We extract fine-grained political topics and bias indicators from weakly labeled news data, addressing the scarcity of fine-grained annotations through an automated topic discovery process. 
  
  \item \textbf{Cross-Encoder Political Bias Embedding:}
  We use mined topics as interpretable dimensions and bias indicators as reference points for bias scoring, developing a weak-label training strategy to enable nuanced political bias comprehension with a political-aware cross-encoder while designing topic retrieval to enhance efficiency.
\end{itemize}

\paragraph{Contributions}
The key contributions of this work are summarized as follows:
\begin{enumerate}[nolistsep,label*=(\arabic*)]
  \item \textbf{Task Formulation and Framework Design:} 
  We introduce the novel task of political bias embedding and present PRISM, the first framework specifically designed to generate interpretable political bias embeddings that go beyond surface-level semantics to capture ideological orientation. 
  
  \item \textbf{Annotation-Free, Interpretable Embedding Approach:}
  PRISM operates without requiring fine-grained manual annotations. 
  It leverages distant supervision to automatically mine controversial political topics and their associated bias indicators. 
  Utilizing a political-aware cross-encoder, PRISM produces embeddings that are inherently interpretable, with dimensions explicitly grounded in semantically meaningful bias indicators.
  
  \item \textbf{Empirical Effectiveness Across Tasks:} 
  Extensive experiments show that PRISM consistently outperforms state-of-the-art baselines on key downstream tasks such as political bias classification and politically diversified content retrieval, while offering transparent insight into ideological representation.
\end{enumerate}

\section{Related Work}
\label{sec:related_work}

\paragraph{Political Bias Mining}
Political bias in news articles has been extensively studied~\citep{baly-etal-2020-written, nakov-etal-2024-survey, martinez-etal-2024-balancing}, with much of the research focused on political bias classification. 
Early work predominantly framed this as a binary classification problem, distinguishing between left-leaning and right-leaning viewpoints~\citep{iyyer-etal-2014-political, chen2017opinion, kulkarni-etal-2018-multi, fan-etal-2019-plain, baly-etal-2020-detect, spinde-etal-2021-neural-media, kim-johnson-2022-close, liu-etal-2022-politics, liu-etal-2023-ideology, hong-etal-2023-disentangling, lin-etal-2024-indivec, liu-etal-2024-encoding}. 

SLAP4SLIP by \citet{hofmann-etal-2022-modeling} focuses on concept discovery and framing analysis, modeling ideological polarization along the dimensions of salience and framing. It employs graph neural networks with structured sparsity to detect polarized concepts without relying on explicit political orientation labels. While related, their objective differs from ours, as SLAP4SLIP emphasizes framing analysis rather than producing interpretable political bias embeddings. Media framing analysis, as surveyed by \citet{otmakhova-etal-2024-media}, offers another lens for understanding political bias by examining how information is "packaged" to elicit specific interpretations, often through emphasis or word choice. Complementary to this, recent work by \citet{sutter-etal-2024-unsupervised} shows that integrating text embeddings with network structures via graph neural networks enhances performance in unsupervised stance detection.

Yet, recognizing the limitations of a single left-to-right scale, particularly in non-U.S. political contexts, researchers have explored multi-dimensional methods that classify texts by ideological stances on specific policy issues (e.g., gun control, abortion)~\citep{kim-johnson-2022-close} or broader ideological dimensions like economic equality and political regime~\cite{liu-etal-2023-ideology, liu-etal-2024-encoding}. 

Beyond classification, politically diversified retrieval systems have been developed to surface content from across the ideological spectrum~\citep{wu-etal-2020-sentirec, draws2021assessing, vrijenhoek2021recommenders, huang2024diversity, sun2024diversinews}. 
However, existing approaches largely rely on categorical labels or metadata rather than embedding-based representations of political bias, limiting their ability to generalize across diverse sources and viewpoints.

\paragraph{Domain-Specific Embedding}
Domain-specific embedding models have proved effective in specialized fields such as biomedical~\citep{chen2019biosentvec, lee2020biobert}, financial~\citep{anderson2024greenback, tang2024we}, and scientific~\citep{beltagy-etal-2019-scibert} domains. 
These models leverage domain-adaptive pretraining to better capture nuances within their respective fields.

For political bias analysis, POLITICS~\cite{liu-etal-2022-politics} fine-tunes RoBERTa~\cite{liu2019roberta} with bias-specific objectives, while recent instruction-following models~\cite{su-etal-2023-one, peng-etal-2024-answer} enable the creation of task-specific embeddings. 
Despite their effectiveness, these models lack interpretability, making it difficult to understand and explain how ideological biases are embedded in their representations~\cite{martinez-etal-2024-balancing}.

\paragraph{Interpretable Text Embedding}
Recent work on interpretable text embeddings either focuses on analyzing existing embeddings~\cite{lee-etal-2022-toward, simhi-markovitch-2023-interpreting} or generating inherently interpretable ones~\cite{opitz-frank-2022-sbert, mcinerney-etal-2023-chill, patel-etal-2023-learning}.
A notable strategy for achieving interpretability using question-answer pairs as embedding dimensions, providing explicit semantic meaning for each dimension~\cite{benara2024crafting, sun2025general}.

However, to our knowledge, no existing work addresses the unique challenges of creating interpretable embeddings for political bias analysis. 
Given the complexity and subjectivity of ideological viewpoints, existing frameworks lack mechanisms to explicitly encode and explain political bias.
Our work bridges this gap by introducing the first framework, PRISM, to produce interpretable political bias embeddings, ensuring both effectiveness and transparency in ideological representation.

\section{The PRISM Framework}
\label{sec:framework}

\begin{figure*}[t]
  \centering
  \captionsetup{skip=0.5em,belowskip=-0.5em}
  \includegraphics[width=0.99\textwidth]{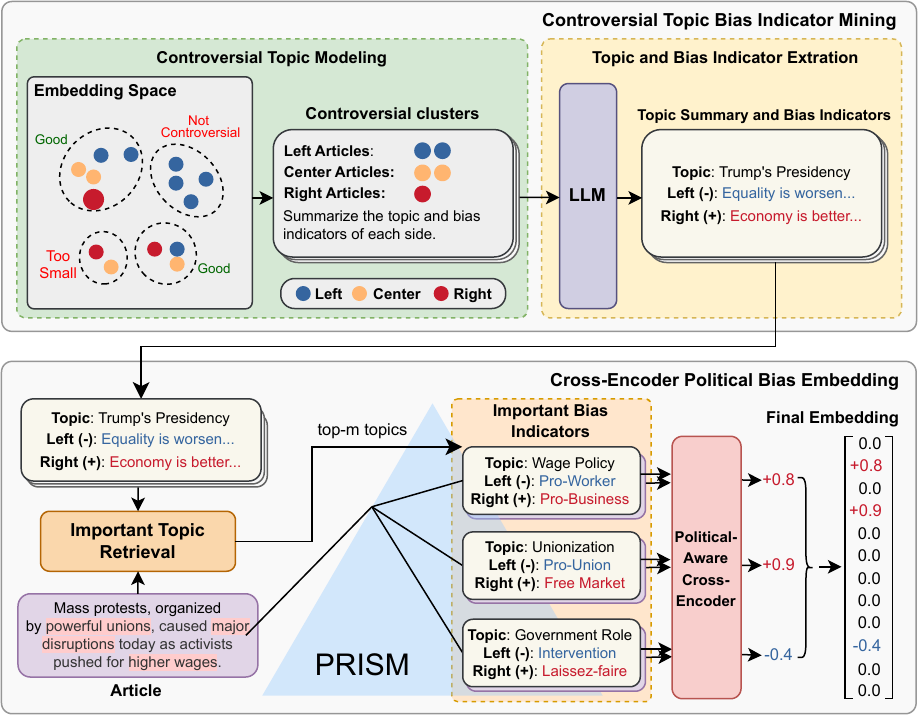}
  \caption{Overview of the PRISM framework.}
  \label{fig:framework}
\end{figure*}

\subsection{Overview}
\label{sec:framework:overview}

We introduce PRISM, a framework designed to generate interpretable political bias embeddings for news articles.
As illustrated in Figure~\ref{fig:framework}, PRISM operates in two key stages: 
\begin{enumerate}[nolistsep,label*=(\arabic*)]
  \item \textbf{Controversial Topic Bias Indicator Mining:} 
  Extracts fine-grained political topics and their corresponding bias indicators from a large, weakly labeled news corpus (Section \ref{sec:framework:topic-mining}). 
  
  \item \textbf{Cross-Encoder Political Bias Embedding:}
  Generates structured embeddings that encode political bias along interpretable dimensions (Section \ref{sec:framework:cross-encoder}).
\end{enumerate}

\paragraph{Mining Political Topics and Bias Indicators}
PRISM first identifies controversial topics and their bias indicators from weakly labeled data. 
For example, a topic extracted from the BigNews dataset~\cite{liu-etal-2022-politics}, \topic{Medicaid Overhaul and Health Care Funding,} has two bias indicators:
\begin{itemize}[nolistsep,left=2pt]
  \item \textbf{Left Indicator:}
  \leftind{Advocates for increased funding for public health, express concern over privatization and potential reductions in services.}
  
  \item \textbf{Right Indicator:}
  \rightind{Focus on cost control, support for block grants and private insurance options, prioritize reducing government spending.}
\end{itemize}

These topics form the embedding dimensions of political bias, while the bias indicators serve as reference points for encoding political bias.

\paragraph{Generating Interpretable Bias Embeddings}
To quantify an article’s stance on each topic, PRISM employs a political-aware cross-encoder model, which assigns a score between 0 and 1 based on how strongly the article aligns with a given bias indicator. 
The resulting embeddings satisfy two essential interpretability properties: 
\begin{itemize}[nolistsep,left=2pt]
  \item \textbf{Selective Activation:} 
  Only relevant and bias-bearing topics receive nonzero values.
  
  \item \textbf{Explicit Bias Representation:} 
  High scores indicate strong alignment with a specific bias, facilitating direct interpretation.
\end{itemize}
Within these two stages, PRISM enhances interpretability while maintaining flexibility across diverse ideological landscapes.
We begin by detailing the controversial topic bias indicator mining stage.

\subsection{Controversial Topic Bias Indicator Mining}
\label{sec:framework:topic-mining}

\paragraph{Motivation}
Capturing political bias requires fine-grained embedding dimensions that reflect ideological differences. 
However, manually curating such dimensions is costly and impractical.
PRISM addresses this challenge by \emph{automatically identifying controversial topics} and their associated \emph{bias indicators} from a weakly labeled news corpus.

\paragraph{Weakly Labeled News Corpus}
PRISM relies on weakly labeled media bias ratings rather than manually annotated data.
Since news outlets often exhibit editorial biases through selective reporting or omission of certain facts~\cite{baly-etal-2020-detect, rodrigo2024systematic}, PRISM utilizes datasets such as NewsSpectrum~\cite{sun2024diversinews} and BigNews~\cite{liu-etal-2022-politics} as our weakly labeled news corpus, where articles are assigned \emph{media bias ratings} (e.g., $-1$ for left, $0$ for center, $1$ for right) based on AllSides,\footnote{\url{https://www.allsides.com/media-bias/ratings}} which provides expert-based bias assessments.

\paragraph{Controversial Topic Modeling}
PRISM identifies controversial topics by leveraging semantic text encoders and clustering techniques:
\begin{enumerate}[nolistsep,label*=(\arabic*)]
  \item \textbf{Encoding the News Corpus:} 
  Each article is transformed into an embedding using a pre-trained semantic text encoder, which maps semantically similar texts to nearby locations in the embedding space.

  \item \textbf{Clustering Similar Articles:} 
  Using $k$-means clustering, articles covering similar topics are grouped together.

  \item \textbf{Measuring Bias Dispersion:} 
  The Bias Dispersion metric quantifies ideological diversity within each cluster by computing the variance of media bias ratings. Clusters with high dispersion indicate controversial topics, as they contain articles from diverse ideological views.
  Formally, for a cluster containing $n$ articles with bias ratings $\bm{R} = \{r_1, r_2, \cdots, r_n\}$, Bias Dispersion is calculated as:
  \begin{equation*}
    \text{Bias Dispersion}(\bm{R}) = \frac{1}{n} \sum_{i=1}^{n} (r_i - \bar{r})^2,
  \end{equation*}
  where $\bar{r}$ is the mean bias rating of the cluster.
  Clusters with a Bias Dispersion exceeding a threshold $\tau$ and containing at least $p$ articles are identified as controversial topics, ensuring they are widely debated and well-represented.
\end{enumerate}

\paragraph{Topic and Bias Indicator Extraction}
For each identified topic, PRISM extracts bias indicators using LLMs.
Specifically, given a set of sample articles and their media bias ratings, the LLM generates: (1) a concise, neutral topic summary and (2) bias indicators describing left-leaning and right-leaning perspectives.
This process allows PRISM to systematically mine topics and define bias indicators without requiring \emph{manual annotation}.
The LLM prompt is provided in Appendix~\ref{app:prompts}.

\subsection{Cross-Encoder Political Bias Embedding}
\label{sec:framework:cross-encoder}

To generate interpretable political bias embeddings, PRISM assigns values to each controversial topic dimension for a given article. This process consists of two pivotal steps:
(1) Important Topic Retrieval, which identifies the most bias-bearing and relevant topics for the given news article;
(2) Political-Aware Cross-Encoder Embedding, which computes bias alignment scores to generate structured embeddings.

\paragraph{Important Topic Retrieval}
A key challenge in bias representation is ensuring that only \emph{relevant} and \emph{bias-bearing} dimensions are assigned nonzero values while filtering out \emph{neutral} or \emph{irrelevant} topics. 
To achieve this, PRISM retrieves the most important topics using pre-trained embeddings of the topics and their corresponding bias indicators.

For each news article, we encode its text, along with the topic summaries and their left and right indicators, using the same semantic text encoder from the previous topic mining stage. 
We then compute an importance score for each topic $i$ using the following equation:
\begin{equation} \label{eqn:topic_importance}
  \text{Score}(i) = \lambda (\bm{x} \cdot \bm{t}_i) + (1-\lambda) |\bm{x} \cdot \bm{r}_i - \bm{x} \cdot \bm{l}_i|,
\end{equation}
where $\bm{x}$ is the embedding of the news article; 
$\bm{t}_i$ is the embedding of topic $i$; 
$\bm{l}_i$ and $\bm{r}_i$ are the embeddings of the left and right indicators of topic $i$; and
$\lambda$ is the weighting factor balancing topic relevance and bias divergence.

Equation \ref{eqn:topic_importance} balances two core factors: 
(1) \emph{Relevance to the topic} ($\bm{x} \cdot \bm{t}_i $) that measures how closely the article aligns with the topic;
(2) \emph{Bias divergence} $|\bm{r}_i \cdot \bm{x} - \bm{l}_i \cdot \bm{x}|$, which captures how strongly the article leans toward one bias over the other.
By selecting the top-$m$ topics with the highest scores, PRISM ensures that embeddings remain efficient and interpretable, focusing only on the most relevant and bias-revealing topics.

\paragraph{Political-Aware Cross-Encoder Embedding}
To ensure that the final embedding selectively activates only for relevant bias dimensions, we develop a new political-aware cross-encoder model that explicitly compares articles with bias indicators rather than relying solely on textual features.

\subparagraph{Training the Cross-Encoder Model}
A cross-encoder model \cite{nogueira2019passage}, parameterized by $\theta$, takes two text inputs and outputs a bias alignment score:
\begin{equation} \label{eqn:bias_align_score}
  f_\theta(\bm{a}, \bm{b}) \in (0,1),
\end{equation}
where $\bm{a}$ denotes a news article; $\bm{b}$ represents the bias indicator (left or right). 
Equation \ref{eqn:bias_align_score} quantifies the alignment between $\bm{a}$ and $\bm{b}$.

\subparagraph{Weak Label Generation for Training}
To train a political-aware cross-encoder, we generate weak supervision labels by leveraging the bias indicators of each topic cluster. 
Given an article $\bm{a}$, we first consider its \emph{in-cluster}'s left $\bm{b}^{\text{left}}$ and right $\bm{b}^{\text{right}}$ indicators and create the following training pairs:
\begin{align*}
  (\bm{a}, \bm{b}^{\text{left}}) &\rightarrow \begin{cases}
    1 & \text{if } bias = \text{left}\\
    0 & \text{otherwise}
  \end{cases},\\
  (\bm{a}, \bm{b}^{\text{right}}) &\rightarrow \begin{cases}
    1 & \text{if } bias = \text{right}\\
    0 & \text{otherwise}
  \end{cases}.
\end{align*}

Additionally, we consider some random \emph{out-of-cluster} topics as negative samples, ensuring that the model does not falsely associate an article with unrelated topics:
\begin{displaymath}
  (\bm{a}, \bm{b}^{\text{left}}) \rightarrow 0,~(\bm{a}, \bm{b}^{\text{right}}) \rightarrow 0.
\end{displaymath}

The model is trained using Mean Squared Error (MSE) loss.
By learning to map article–indicator pairs to these labels, it is expected that the model can accurately distinguish relevant biases from neutral content and focus on topic-specific bias signals rather than generic political leanings.

\subparagraph{Generating the Final Bias Embedding}
During inference, PRISM produces the final embedding vector by computing bias alignment scores for the top-$m$ important topics.

Specifically, given a trained cross-encoder $f_{\theta}$, we generate the bias embedding as follows.
First, we initialize the embedding vector with a list of zero values. 
Then, we retrieve the top $m$ most important topics $\bm{M}$ for the article $\bm{a}$. 
Lastly, we compute bias alignment scores using the cross-encoder for each selected topic. For each topic $i \in \bm{M}$, the final embedding value $e_i$ is computed as:
\begin{equation}\label{eqn:final_embedding}
  e_i =
  \begin{cases}
    s^{r}_{i} - s^{l}_{i}, & \text{if}~i \in \bm{M},\\
    0,   & \text{otherwise}.
  \end{cases}
\end{equation}
where $s^{l}_{i} = f_{\theta}(\bm{a}, \bm{b}^{\text{left}}_{i})$ and $s^{r}_{i} = f_{\theta}(\bm{a}, \bm{b}^{\text{right}}_{i})$ are the bias alignment scores with the left indicator $\bm{b}^{\text{left}}_{i}$ and the right indicator $\bm{b}^{\text{right}}_{i}$. 

\subparagraph{Remarks}
In Equation \ref{eqn:final_embedding}, PRISM encodes political bias by computing the difference between left and right bias scores, ensuring: (1) Positive values indicate a right-leaning bias; (2) Negative values indicate a left-leaning bias; and (3) Zero values indicate neutrality or irrelevance.
This cross-encoder embedding approach guarantees that PRISM's embeddings remain interpretable, efficient, and explicitly tied to bias-revealing dimensions, validated through empirical analysis in Section \ref{sec:expt}.

\section{Experiments}
\label{sec:expt}

We evaluate the effectiveness of PRISM by addressing the following key research questions:
\begin{itemize}[nolistsep,left=2pt]
  \item \textbf{RQ1 (Political Bias Signal Quality):} 
  How well does PRISM capture bias-related signals compared to generic semantic text embeddings and political bias-specific models? (Section \ref{sec:exp:cls})
  
  \item \textbf{RQ2 (Distance Measurement Effectiveness):} 
  To what extent does the political bias embedding space accurately reflect ideological similarities between articles, making it suitable for diversified retrieval? (Section \ref{sec:exp:retrieval})
  
  \item \textbf{RQ3 (Interpretability):} 
  Can the political bias embedding provide meaningful insights that users can intuitively interpret? (Section \ref{sec:exp:case-study})
\end{itemize}

Before presenting results, we outline the datasets and benchmark models used for evaluation.

\subsection{Datasets}
\label{sec:expt:datasets}

We conduct experiments on two large-scale, real-world news datasets that provide extensive political coverage and diverse ideological perspectives.
\begin{itemize}[nolistsep,left=2pt]
  \item \textbf{NewsSpectrum} \cite{sun2024diversinews} consists of 250,000 news articles sourced from 961 distinct media outlets, with each article assigned a bias score ranging from left (-2) to right (2). 
  This dataset is carefully curated to maintain a balanced distribution of political perspectives, making it particularly useful for evaluating models in diverse ideological settings.
  
  \item \textbf{BigNews} \cite{liu-etal-2022-politics} comprises 3.6 million news articles collected from 13 media outlets, each labeled with its corresponding media bias. 
  This large-scale dataset provides broad coverage of political discourse across various events and ideological stances.
\end{itemize}

\subsection{Benchmark Models}
\label{sec:expt:models}

Since PRISM is the first framework designed to produce interpretable political bias embeddings, there is no direct competitor. To systematically assess its performance, we compare it against state-of-the-art models from four relevant areas:
\begin{itemize}[nolistsep,left=2pt]
  \item \textbf{Generic Semantic Text Embedding Models:}
  We select \textbf{AnglE} (\chkp{UAE-Large-V1}) \cite{li-li-2024-aoe}, a state-of-the-art text embedding model widely used for various NLP tasks \cite{muennighoff-etal-2023-mteb}, as a strong baseline for generic semantic embeddings.

  \item \textbf{Political Bias-Specific Models:}
  We include \textbf{POLITICS}~\cite{liu-etal-2022-politics}, the leading model for political bias analysis, pre-trained on BigNews. 
  For evaluation, we use the official Hugging Face checkpoint \chkp{launch/POLITICS} and extract embeddings from the \chkp{CLS} token's last hidden state.

  \item \textbf{Instruction-Following Embedding Models:}
  We evaluate two cutting-edge instruction-following embedding models: \textbf{InstructOR} (\chkp{instructor-large}) \cite{su-etal-2023-one} and \textbf{InBedder} (\chkp{roberta-large-InBedder}) \cite{peng-etal-2024-answer}, provided with specific instructions for political bias analysis. 

  \item \textbf{Interpretable Text Embedding Models:}
  To evaluate interpretability, we compare against \textbf{CQG-MBQA} \cite{sun2025general}, the state-of-the-art interpretable text embedding model. We use its publicly pre-trained checkpoint.\footnote{\url{https://github.com/dukesun99/CQG-MBQA}}
\end{itemize}

\begin{table*}[t]
\renewcommand{\arraystretch}{1.2}
\small
\centering
\begin{tabular}{cccccccccccc}
  \toprule
  \rowcolor[HTML]{FFF2CC}
  \multirow{2.5}{*}{\cellcolor[HTML]{FFF2CC}\textbf{Model}} & \multicolumn{5}{c}{\textbf{NewsSepctrum}} & \multicolumn{5}{c}{\textbf{BigNews}} \\ 
  \cmidrule(lr){2-6} \cmidrule(lr){7-11}
  \rowcolor[HTML]{FFF2CC}
  & \textbf{Acc $\uparrow$} & \textbf{Pre $\uparrow$} & \textbf{Rec $\uparrow$} & \textbf{F1-Ma $\uparrow$} & \textbf{F1-Mi $\uparrow$} & \textbf{Acc $\uparrow$} & \textbf{Pre $\uparrow$} & \textbf{Rec $\uparrow$} & \textbf{F1-Ma $\uparrow$} & \textbf{F1-Mi $\uparrow$} \\
  \midrule
  \textbf{AnglE}      & 48.4 & 48.2 & 48.8 & 48.2 & 48.4 & 69.0 & 69.0 & 69.0 & 69.0 & 69.0 \\ 
  \textbf{Instructor} & 47.9 & 47.8 & 48.5 & 47.7 & 48.0 & 63.7 & 63.7 & 63.7 & 63.7 & 63.7 \\
  \textbf{InBedder}   & 50.2 & 50.2 & 50.8 & 49.8 & 50.2 & 64.6 & 64.6 & 64.6 & 64.6 & 64.6 \\
  \textbf{CQG-MBQA}   & 45.1 & 44.9 & 45.5 & 44.9 & 45.1 & 61.0 & 61.0 & 61.0 & 61.0 & 61.0 \\
  \textbf{POLITICS}   & 51.3 & 51.7 & 51.9 & 51.1 & 51.3 & \textbf{85.7} & \textbf{85.7} & \textbf{85.7} & \textbf{85.7} & \textbf{85.7} \\
  \rowcolor[HTML]{D9EAD3}
  \textbf{PRISM}      & \textbf{86.1} & \textbf{86.5} & \textbf{86.2} & \textbf{86.2} & \textbf{86.1} & 73.5 & 81.3 & 73.7 & 74.0 & 73.5 \\
  \bottomrule
\end{tabular}
\caption{Political bias classification results on NewsSpectrum and BigNews, evaluated using Accuracy (Acc), Precision-Macro (Pre), Recall-Macro (Rec), F1-Macro (F1-Ma), and F1-Micro (F1-Mi).}
\label{tab:political-bias-cls}
\vspace{0.25em}
\end{table*}

\subsection{Political Bias Classification}
\label{sec:exp:cls}

To assess how effectively PRISM captures political bias in news articles, we evaluate its performance on a political bias classification task.

\paragraph{Experimental Setup}
We train an SVM classifier using embedding vectors generated by different models.
Training is performed on a held-out dataset, distinct from PRISM's training data, and performance is evaluated on a separate test set. 
The classification results are presented in Table \ref{tab:political-bias-cls}.

\paragraph{Result Analysis}
As shown in Table~\ref{tab:political-bias-cls}, PRISM outperforms all baseline models, achieving the highest classification accuracy on NewsSpectrum and the second highest on BigNews.
These results demonstrate PRISM's ability to effectively capture political bias signals while maintaining interpretability.
Unlike generic semantic embeddings such as AnglE, which primarily encode overall content similarity, PRISM explicitly models ideological orientations, leading to superior performance.

Notably, while POLITICS achieves higher accuracy than PRISM on BigNews, this advantage stems from the test set of BigNews being part of its training data. 
Yet, when evaluated on NewsSpectrum, which was not seen during training, POLITICS lags significantly behind PRISM.
This suggests that PRISM generalizes better across different datasets, reinforcing its ability to capture ideological bias without overfitting to specific corpora.

Overall, this experiment demonstrates that PRISM not only produces interpretable embeddings but also retains strong predictive power, establishing it as a robust and generalizable framework for political bias analysis.

\subsection{Diversified Retrieval}
\label{sec:exp:retrieval}

A fundamental characteristic of political bias embeddings is their ability to serve as distance metrics for ideological similarity, making them particularly valuable for retrieval tasks.
To evaluate this capability, we conduct politically diversified retrieval experiments, following the DiversiNews framework~\citep{sun2024diversinews}.

\begin{figure}[t]
  \centering%
  \captionsetup{skip=0.5em,belowskip=0em}%
  \includegraphics[width=0.99\columnwidth]{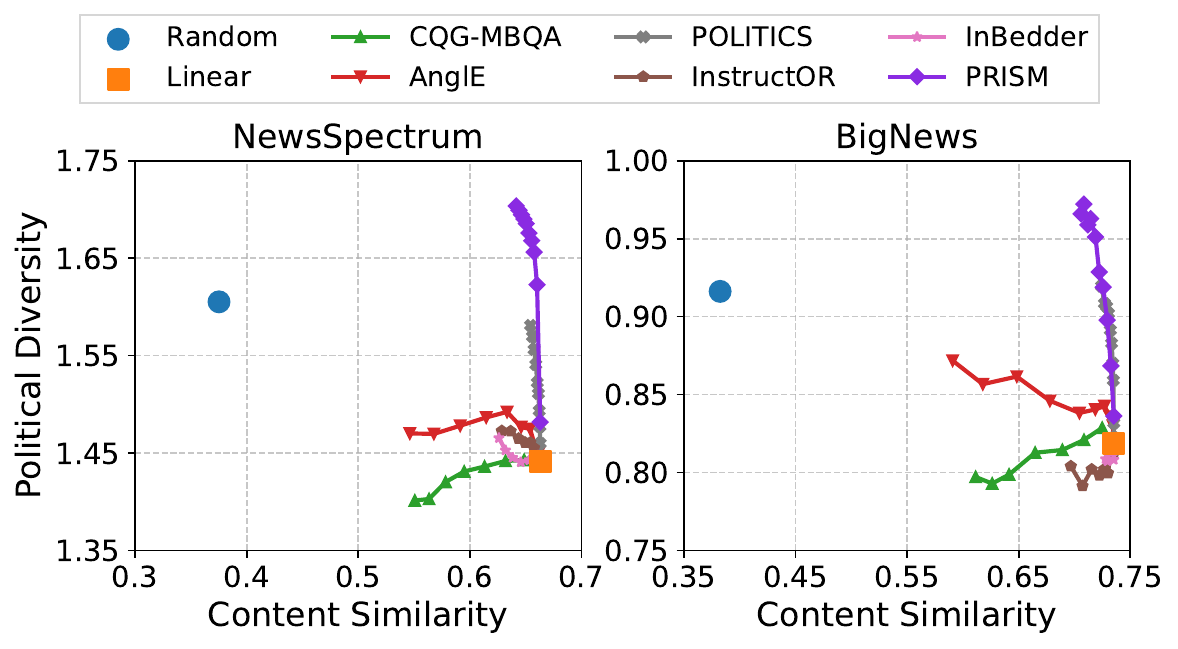}
  \caption{Diversified retrieval results.}%
  \label{fig:diversified-retrieval}
\end{figure}

\paragraph{Experimental Setup}
We adopt the retrieval protocol and evaluation metrics from DiversiNews~\citep{sun2024diversinews} and employ the Diversity-aware $k$-Maximum Inner Product Search (D$k$MIPS) algorithm \cite{huang2024diversity} to enhance political diversity in retrieval results.
Our implementation incorporates two distinct embedding spaces: 
\begin{itemize}[nolistsep,left=2pt] 
  \item \textbf{AnglE Embeddings:} Used to measure query-document relevance based on inner product similarity. 
  
  \item \textbf{Model-specific Embeddings} (e.g., PRISM, POLITICS, InstructOR): Used to quantify inter-document political diversity based on ideological differences. 
\end{itemize}
This dual-space design enables us to systematically evaluate each model's capacity to encode and differentiate political bias in retrieval scenarios.

\begin{figure*}[t]
  \centering
  \captionsetup{skip=0.5em,belowskip=-0.5em}
  \includegraphics[width=0.99\textwidth]{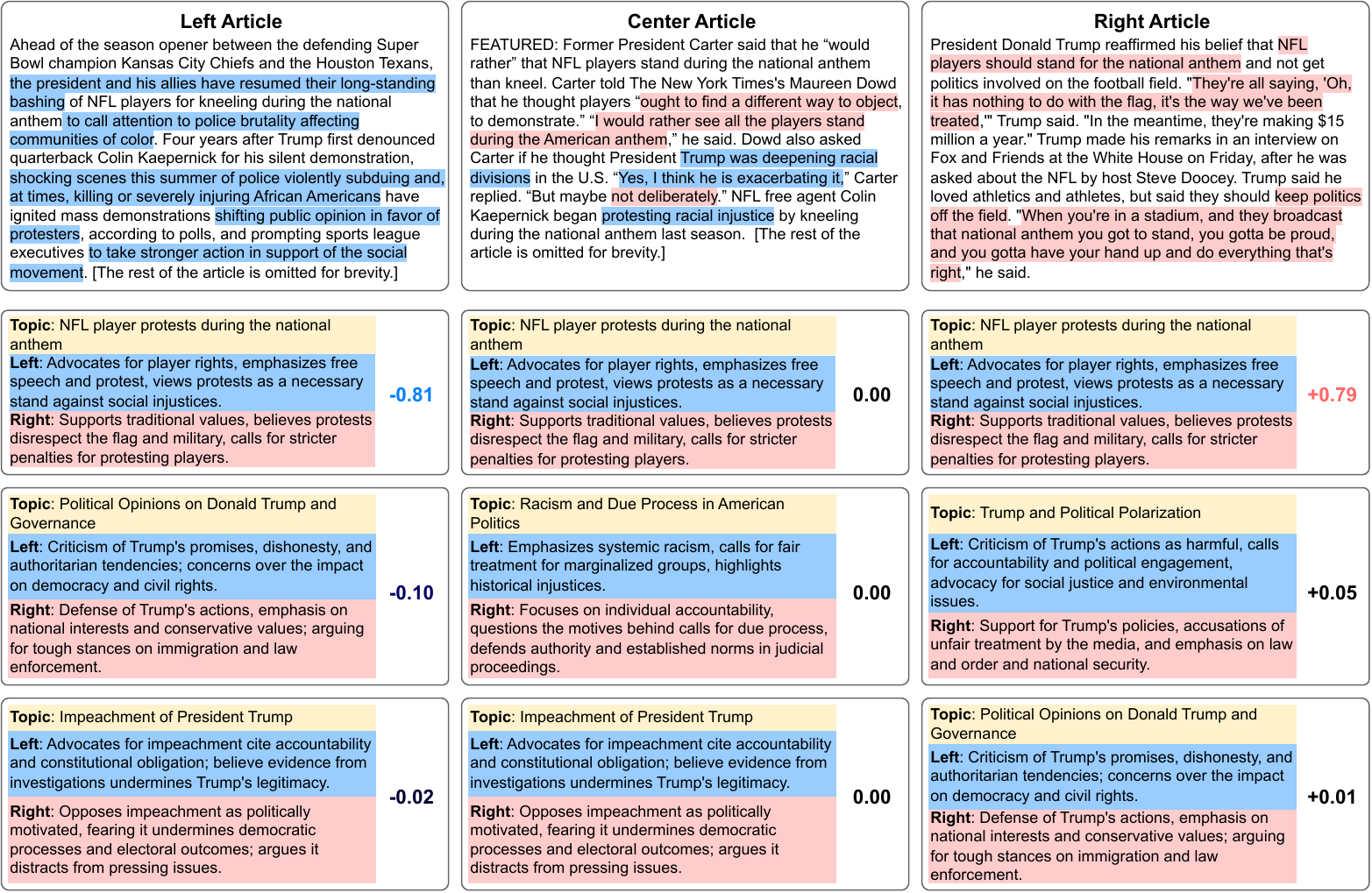}
  \caption{Case study on three news articles with different political bias.}
  \label{fig:case-study}
  \vspace{-0.25em}
\end{figure*}

We evaluate retrieval performance using the following measures:
\begin{itemize}[nolistsep,left=2pt]
  \item \textbf{Content Similarity:} 
  Given a retrieved set $\bm{S}=\{\bm{p}1, \bm{p}_2, \cdots, \bm{p}_k\}$ that contains $k$ news articles for a query $\bm{q}$, the content similarity is defined as the mean inner product between each retrieved article $\bm{p}_i$ and the query $\bm{q}$:
  \begin{equation*}
    \text{Sim}(\bm{S}, \bm{q}) = \frac{1}{k} \sum_{i=1}^{k} \ip{\bm{p}_i, \bm{q}},
  \end{equation*}
  where $\ip{\bm{p}_i, \bm{q}}$ represents the inner product similarity between article $\bm{p}_i$ and query $\bm{q}$. 
  Higher values indicate stronger content relevance.
    
  \item \textbf{Political Diversity:} 
  To measure the political diversity of retrieved results, we compute the mean pairwise difference between the bias ratings $r_i$ of articles in $\bm{S}$:
  \begin{equation*}
    \text{Div}(\bm{S}) = \frac{2}{k(k-1)} \sum_{i=1}^{k-1} \sum_{j=i+1}^{k} |r_i - r_j|.
  \end{equation*}
  Higher values indicate a greater spread of political perspectives within the retrieved articles $\bm{S}$, ensuring ideological balance in the results.
\end{itemize}

\paragraph{Result Analysis}
Figure~\ref{fig:diversified-retrieval} illustrates the trade-off between content similarity and political diversity across different retrieval methods. 
The results show that PRISM consistently outperforms all baselines in both dimensions:
(1) At equivalent levels of political diversity, PRISM preserves higher content relevance than competing models;
(2) At comparable content similarity, it delivers greater ideological diversity in the retrieved results.

This consistent advantage across both datasets highlights two key strengths of PRISM:
\begin{itemize}[nolistsep,left=2pt]
  \item \textbf{Better Bias Representation:} PRISM's embedding space more effectively captures ideological relationships between news articles compared to existing approaches.

  \item \textbf{Reliable Distance Metric:} The embeddings produced by PRISM serve as an effective metric for politically diversified retrieval, balancing relevance and ideological diversity.
\end{itemize}

\begin{figure*}[t]
  \centering
  \captionsetup{skip=0.5em,belowskip=0em}
  \includegraphics[width=0.9\textwidth]{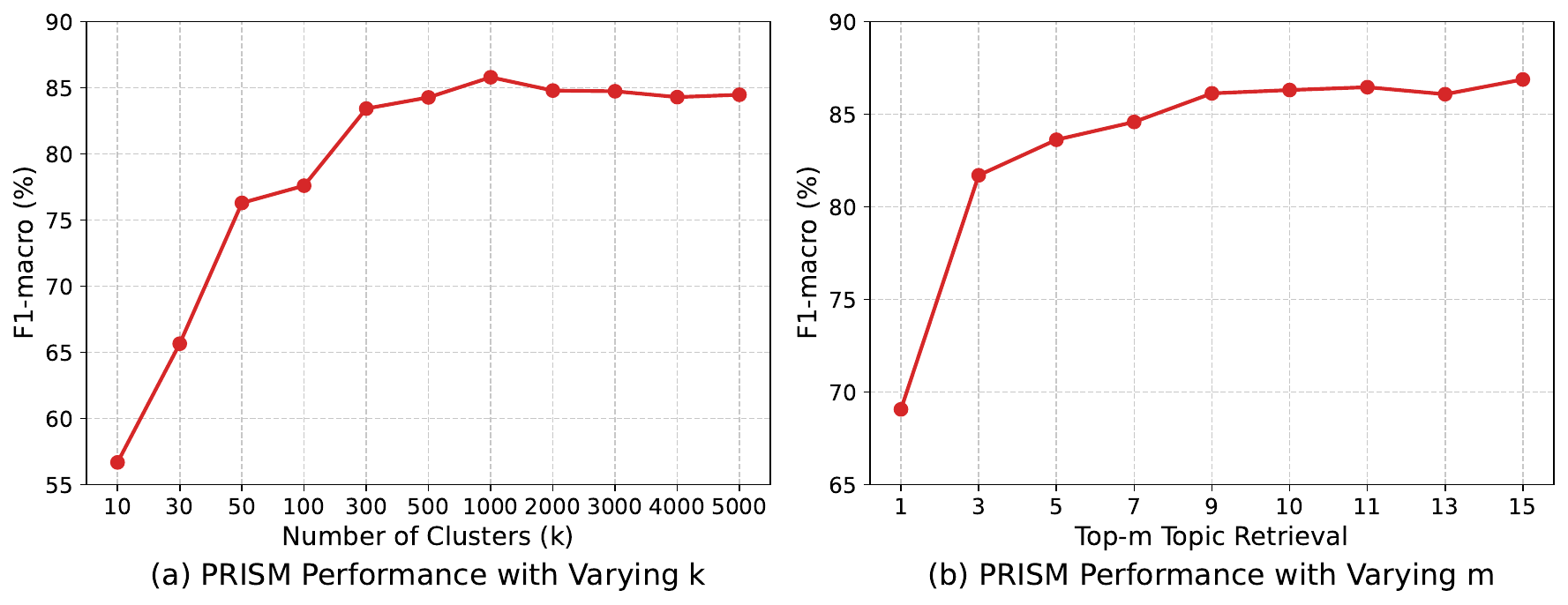}
  \caption{Effect of key parameters on classification performance (F1-marco) on NewsSpectrum: (a) number of clusters $k$ for topic mining; (b) number of top-$m$ topic dimensions retrieved per article.}
  \label{fig:parameter-study}
\end{figure*}

\subsection{Case Study}
\label{sec:exp:case-study}

To illustrate PRISM's interpretability, we present a detailed case study analyzing its embedding representations for three news articles with different political orientations (Figure \ref{fig:case-study}).

\paragraph{Overview}
The selected articles examine the ``\textbf{NFL player protests during the national anthem (2015-2020)},'' a politically charged topic with clear ideological divides. 
Each article is sourced from media outlets with distinct political leanings, reflecting contrasting bias indicators:
\begin{itemize}[nolistsep,left=2pt]
  \item \textbf{Left-leaning:} Highlights player rights, freedom of expression, and social justice.
  \item \textbf{Right-leaning:} Focuses on traditional values, patriotism, and respect for national symbols.
\end{itemize}
To aid visual interpretation, opinion-bearing text is color-coded: blue for left-leaning indicators and red for right-leaning ones.

\paragraph{Findings and Interpretability Analysis}
PRISM exhibits several key capabilities that enhance its interpretability and efficacy in bias representation: 
\begin{itemize}[nolistsep,left=2pt]
  \item \textbf{Accurate Topic Identification:} 
  PRISM accurately identifies NFL player protests as the primary topic and assigns bias scores aligned with each article's ideological stance (-1 for left, 0 for center, and 1 for right).
  
  \item \textbf{Nuanced Topic Weighting:} 
  Related topics receive proportionally smaller values, reflecting their secondary relevance. Unrelated topics are assigned values close to zero, ensuring embedding sparsity and interpretability.
  
  \item \textbf{Clear and Intuitive Bias Representation:} 
  Users can interpret embeddings via bias indicators, while sparsity ensures only relevant dimensions hold meaningful values, minimizing noise and enhancing clarity.
\end{itemize}

\subsection{Parameter Study}
\label{sec:exp:param}

We conduct a parameter study to analyze the sensitivity effects of two key components in PRISM: (1) the number of clusters $k$ used in topic mining and (2) the number of top-$m$ topic dimensions retrieved per article during classification. 
All experiments are conducted on the NewsSpectrum dataset, with results summarized in Figure~\ref{fig:parameter-study}.

\paragraph{Effect of Number of Clusters ($k$)}
We vary the number of clusters $k$ used in the $k$-means topic mining stage from 10 to 5,000, without re-training the cross-encoder model. 
To ensure fair comparison, we proportionally adjust the minimum cluster size ($p$). 
As shown in Figure~\ref{fig:parameter-study}(a), performance improves steadily as $k$ increases, peaking around $k=1{,}000$, after which it begins to decline slightly. 
This highlights the importance of choosing an appropriate number of clusters: too few clusters limit topic diversity, while too many may introduce noise or redundancy in the topic space.

\paragraph{Effect of Top-$m$ Topic Retrieval ($m$)}  
We also evaluate the influence of the number of top-$m$ topic dimensions retrieved for each article. 
As illustrated in Figure~\ref{fig:parameter-study}(b), increasing $m$ leads to improved classification performance, with the F1-macro score rising steadily from $m=1$ to $m=9$, and stabilizing beyond that. 
This suggests that retrieving multiple relevant ideological dimensions enriches the representation, while further expansion offers diminishing returns.
\section{Conclusions}
\label{sec:conclusions}

In this work, we introduce Political Bias Embedding, a new task aimed at representing ideological orientations in a structured and interpretable manner.
We propose PRISM, the first framework designed to produce interpretable political bias embeddings. 
By integrating automated topic mining with a new political-aware cross-encoder embedding approach, PRISM effectively captures political bias while maintaining interpretability.
Extensive experiments demonstrate PRISM's superiority over existing models in political bias classification and diversified retrieval. 
Unlike standard semantic embeddings, PRISM encodes ideological distinctions while offering transparent bias insights, making it ideal for bias-aware retrieval and analysis.

\section*{Limitations}
\label{sec:limitations}

Despite PRISM's strengths in robustness and interpretability for political bias embedding, several limitations remain:

\paragraph{Efficiency and Scalability}  
Although PRISM delivers strong performance, its two-stage design--consisting of topic mining and cross-encoder-based embedding--incurs higher computational cost compared to standard embedding models. 
This trade-off is justified by its substantial gains in interpretability and predictive accuracy.
Nevertheless, future work could explore more efficient alternatives, such as model distillation and lightweight encoders, to improve scalability and reduce inference latency without sacrificing performance.

\paragraph{Topic Granularity and Representation Assumptions}  
PRISM relies on clustering-based topic extraction, which necessitates careful tuning of the number of clusters. 
Too few clusters risk oversimplifying ideological nuance, while too many may introduce noise or lead to overly sparse embeddings.
Moreover, PRISM treats each topic dimension as an independent axis in the embedding space, implicitly assuming orthogonality among topics. 
While the use of top-$m$ topic retrieval allows for soft activation across multiple dimensions, this representation may still overlook inter-topic dependencies and correlated ideological dimensions.
Future work could explore more expressive embedding structures that capture semantic overlap or hierarchical relationships among topics.

Furthermore, the current framework may conflate topic and stance, as fine-grained clusters often reflect both thematic content and ideological framing. 
Explicit disentanglement of these two elements, potentially through multi-view representation learning or factorized embeddings, could enhance the interpretability and generalizability of the learned representations.

\paragraph{Evaluation Across Languages, Cultures, and Time}  
Our current evaluation is limited to English-language news from U.S.-based media, annotated using AllSides bias ratings. 
Extending PRISM to other linguistic and cultural contexts would require region-specific corpora and localized bias references, as ideological dimensions may vary significantly across geographies.
While PRISM is designed to be extensible, which enables new political topics to be incorporated via re-running the topic mining process on updated corpora, we do not explicitly evaluate its temporal robustness or responsiveness to emerging discourse. 
Future work may investigate time-sensitive evaluations to assess how well PRISM adapts to shifting ideological landscapes, including the emergence of new topics or changes in framing over time.

\section*{Acknowledgments}
Yiqun Sun and Anthony T. H. Tung were supported by the Ministry of Education, Singapore, under its MOE AcRF TIER 3 Grant (MOE-MOET32022-0001), and by the National Research Foundation, Singapore, under its AI Singapore Programme (AISG Award No.~AISG3-RP-2022-029).
Qiang Huang and Jun Yu were supported by the National Natural Science Foundation of China under grant Nos.~62125201 and U24B20174.
Any opinions, findings, and conclusions or recommendations expressed in this material are those of the author(s) and do not reflect the views of the Ministry of Education, Singapore, or the National Research Foundation, Singapore.

\bibliography{acl_main}

\appendix

\section{Prompts}
\label{app:prompts}

The following prompt is used to mine controversial topics and their associated bias indicators from news articles. 
For each cluster of articles, we provide the article texts and their corresponding media bias labels. 
The prompt instructs the language model to:
\begin{enumerate}[nolistsep,label*=(\arabic*)]
  \item Identify a common topic that reflects the primary point of contention
  
  \item Summarize the topic neutrally
  
  \item Extract distinct bias indicators for both left and right political perspectives
  
  \item Output the extracted topics and bias indicators in a structured format
\end{enumerate}

This prompt design ensures consistent, structured extraction of topics and their associated political perspectives while maintaining neutrality in topic descriptions.

\begin{tcolorbox}
Please summarize the following texts into a common topic, which the VAST MAJORITY of the texts debate on, and can reflect the bias of the texts, which different biases (left, center, right) hold different views on this topic. 

Note that there are multiple sides to the topic. Please summarize the topic in a neutral tone. Please return the topic and the bias indicators, without any other words or sentences. Please summarize the topic in fewer than 10 words. 
\\\\
Give me the results in the following format: Topic: $\langle$Topic$\rangle$

Left Indicator: $\langle$Some key points that Left or Lean Left have$\rangle$

Right Indicator: $\langle$Some key points that Right or Lean Right have$\rangle$
\\\\
Text: \{News\_Article\_i\}

Bias: \{Media\_Bias\_of\_News\_Article\_i\}
\\\\
\{Remaining Texts and Biases Omitted\}

...\\
...
\end{tcolorbox}

\section{Implementation Details}
\label{app:details}

Table \ref{tab:hyperparameters} summarizes the hyperparameters used in the experiments.

\paragraph{Controversial Topic Bias Indicator Mining}
We use \chkp{UAE-Large-V1} as the pre-trained encoder and apply $k$-means clustering with $k = 3{,}000$ to identify topic clusters. 
During the cluster filtering step, we set dataset-specific thresholds: a Bias Dispersion threshold of $\tau = 1.0$ for NewsSpectrum and $\tau = 0.5$ for BigNews, with a minimum cluster size of $p = 50$ for both. 
These thresholds are calibrated to account for the different label granularities in each dataset. 
For topic summarization and bias indicator generation, we randomly sample 50 texts per prompt and use \chkp{GPT-4o-mini} as the language model. 
This process yields 1,810 topics for NewsSpectrum and 2,279 topics for BigNews.

\paragraph{Cross-Encoder Political Bias Embedding}
In the important topic retrieval component, we set the weighting factor $\lambda=0.8$ to balance topic relevance and bias divergence. 
The political-aware cross-encoder model is implemented using \chkp{microsoft/deberta-v3-large} (304M) \cite{he2023debertav3}, trained with weak labels using a learning rate $\alpha=10^{-6}$ and batch size $b=4$. 
Training continues until loss convergence, which occurs at approximately 1,900,000 steps.

\paragraph{Political Bias Classification}
We maintain the original label taxonomies for both datasets: a five-point scale \{-2, -1, 0, 1, 2\} for NewsSpectrum and a three-point scale \{-1, 0, 1\} for BigNews. 
We randomly sample 10,000 articles from NewsSpectrum and 100,000 articles from BigNews. 
Using scikit-learn version 1.5.2, we train an SVM classifier with default parameters\footnote{\url{https://scikit-learn.org/1.5/modules/generated/sklearn.svm.SVC.html}} on 90\% of the evaluation partition and test on the remaining 10\%. 
All results are reported from a single experimental run. 

\begin{table}[t]
\centering
\setlength\tabcolsep{4pt}
\renewcommand{\arraystretch}{1.3}
\resizebox{1.0\columnwidth}{!}{
\begin{tabular}{L{0.36\textwidth} ll}
  \toprule
  \rowcolor[HTML]{FFF2CC}
  \textbf{Description} & \textbf{Symbol} & \textbf{Setting} \\
  \midrule
  Number of clusters & $k$ & 3,000 \\
  Bias dispersion threshold (NewsSpectrum) & $\tau$ & 1.0 \\
  Bias dispersion threshold (BigNews) & $\tau$ & 0.5 \\
  Weighting factor for important topic retrieval & $\lambda$ & 0.8 \\
  Minimum cluster size & $p$ & 50 \\
  Number of topics (NewsSpectrum) & $|M|$ & 1,810 \\
  Number of topics (BigNews) & $|M|$ & 2,279 \\
  Learning rate & $\alpha$ & $10^{-6}$ \\
  Batch size & $b$ & 4 \\
  \bottomrule
\end{tabular}}
\caption{Hyperparameters used in our experiments.}
\label{tab:hyperparameters}
\vspace{-0.5em}
\end{table}

\paragraph{Diversified Retrieval}
For the politically diversified retrieval experiment, we employ an extended version of the BC-Greedy-Avg algorithm of D$k$MIPS~\cite{huang2024diversity} that operates in dual embedding spaces. The objective function is formulated as:
\begin{equation*}
f(\bm{S}) = \frac{\lambda}{k} \sum_{i \in \bm{S}} \ip{\bm{p_i}, \bm{q}} - \frac{2\mu(1-\lambda)}{k(k-1)} \sum_{i \neq j \in \bm{S}} \ip{\hat{\bm{p}}_i, \hat{\bm{p}}_j},
\end{equation*}
where $\bm{S}$ denotes the result set, $\bm{q}$ represents the query vector, $\bm{p}_i$ is the $i$-th document vector in the first space measuring query-document similarity, and $\hat{\bm{p}}_i$ represents the $i$-th document vector in the second space quantifying political similarity between candidates. 

In our implementation, we utilize AnglE-generated embeddings for the first space to measure query-document relevance, while the second space employs method-specific embeddings to capture political diversity. 
The trade-off between diversity and relevancy is controlled by the hyperparameter $\mu \in (0, 1)$, with $\lambda$ fixed at 0.5 across all experiments. 
The implementation of our modified D$k$MIPS algorithm is available at \url{https://github.com/dukesun99/PyDkMIPS}.

\begin{table*}[t]
\centering
\small
\renewcommand{\arraystretch}{1.5}
\setlength{\tabcolsep}{4pt}
\resizebox{0.99\textwidth}{!}{
\begin{tabular}{L{0.18\textwidth} L{0.41\textwidth} L{0.41\textwidth}}
  \toprule
  \rowcolor[HTML]{FFF2CC}
  \textbf{Generated Topic} & \textbf{Left Indicator} & \textbf{Right Indicator} \\
  \midrule
  U.S. Funding and Relationship with the WHO & Criticism of U.S. withdrawal, emphasis on global cooperation; concerns that defunding undermines pandemic response; calls for accountability from WHO without cutting ties. & WHO accused of being ``China-centric,'' calls for defunding until reforms are made; WHO's handling of the pandemic criticized as ineffective; demands for the resignation of WHO leadership. \\ 
  \midrule
  \rowcolor[HTML]{DDEBFF}
  Disappearance and Violence in Mexico & Emphasis on femicide, human rights violations, and environmental activism; criticizes government inaction and suggests increasing violence against marginalized groups. & Focus on cartel violence and law enforcement failures; assertion of a need for stronger government action against organized crime and support for pro-family values amidst crisis. \\ 
  \midrule
  Driver's License Regulations and Data Privacy & Emphasis on civil rights, privacy concerns over data selling by DMVs, criticism of government surveillance, and policies that support undocumented immigrants. & Focus on national security, criticism of relaxed immigration and driving laws for undocumented individuals, and concerns about the criminal misuse of counterfeit IDs. \\ 
  \midrule
  \rowcolor[HTML]{DDEBFF}
  Electric Vehicles and Alternative Powertrains & Advocates for government support, stricter regulations on emissions, and innovation in sustainable technology. & Emphasizes market-driven solutions, skepticism about government overreach and regulations, and preference for traditional vehicles. \\ 
  \bottomrule
\end{tabular}}
\caption{Examples of generated topics and bias indicators.}
\label{tab:bias-examples}
\vspace{-0.5em}
\end{table*}

\section{LLM Generation Quality}
\label{app:quality}

To further illustrate the quality of the generated topics, we include four examples in Table \ref{tab:bias-examples}. 
These topic dimensions exhibit clear, ideologically coherent indicators aligned with widely recognized partisan perspectives.

\section{Additional Experiments}
\label{app:extra-expts}

To further evaluate the generalizability and robustness of PRISM, we conduct three additional experiments: (1) classification on a human-annotated dataset (BASIL), (2) an alignment of label schemes for consistent cross-dataset comparison, and (3) a comparison against zero-shot LLM baselines.

\begin{table}[t]
\centering
\small
\setlength\tabcolsep{4pt}
\renewcommand{\arraystretch}{1.3}
\resizebox{0.99\columnwidth}{!}{
\begin{tabular}{cccccc}
  \toprule
  \rowcolor[HTML]{FFF2CC}
  \textbf{Model} & \textbf{Acc $\uparrow$} & \textbf{Pre $\uparrow$} & \textbf{Rec $\uparrow$} & \textbf{F1-Ma $\uparrow$} & \textbf{F1-Mi $\uparrow$} \\
  \midrule
  \textbf{AnglE}      & 35.0 & 28.8 & 28.9 & 28.8 & 34.7 \\ 
  \textbf{Instructor} & 31.7 & 25.0 & 25.6 & 25.1 & 30.4 \\ 
  \textbf{InBedder}   & 33.3 & 29.0 & 26.7 & 26.8 & 32.8 \\ 
  \textbf{CQG-MBQA}   & 30.0 & 19.2 & 21.1 & 20.0 & 28.3 \\ 
  \textbf{POLITICS}   & 31.7 & 28.9 & 28.9 & 28.9 & 31.5 \\ 
  \rowcolor[HTML]{D9EAD3}
  \textbf{PRISM}      & \textbf{40.0} & \textbf{38.5} & \textbf{36.7} & \textbf{37.3} & \textbf{39.7} \\ 
  \bottomrule
\end{tabular}}
\vspace{-0.5em}
\caption{Classification results on BASIL.}
\label{tab:basil}
\end{table}

\paragraph{Classification on BASIL}
We evaluate PRISM on the BASIL dataset \cite{fan-etal-2019-plain}, a human-annotated corpus that labels each article as \textit{Liberal}, \textit{Center}, or \textit{Conservative}.
Notably, BASIL is entirely disjoint from any data used to train PRISM or its baselines, offering a strong test of out-of-distribution generalization.

For this experiment, we train PRISM on a mixture of the BigNews and NewsSpectrum datasets, then apply a logistic regression classifier to predict the political stance of BASIL articles using PRISM embeddings.
As shown in Table~\ref{tab:basil}, PRISM achieves the highest classification performance among all compared models, including POLITICS. 
These results further highlight the generalizability and robustness of our PRISM framework.

\begin{table}[t]
\centering
\small
\setlength\tabcolsep{4pt}
\renewcommand{\arraystretch}{1.3}
\resizebox{0.99\columnwidth}{!}{
\begin{tabular}{cccccc}
  \toprule
  \rowcolor[HTML]{FFF2CC}
  \textbf{Model} & \textbf{Acc $\uparrow$} & \textbf{Pre $\uparrow$} & \textbf{Rec $\uparrow$} & \textbf{F1-Ma $\uparrow$} & \textbf{F1-Mi $\uparrow$} \\
  \midrule
  \textbf{AnglE}      & 62.7 & 61.6 & 58.3 & 58.9 & 62.7 \\ 
  \textbf{Instructor} & 60.4 & 59.5 & 55.4 & 55.9 & 60.4 \\ 
  \textbf{InBedder}   & 64.1 & 66.7 & 58.3 & 59.0 & 64.1 \\ 
  \textbf{CQG-MBQA}   & 56.0 & 55.7 & 51.6 & 52.2 & 56.0 \\ 
  \textbf{POLITICS}   & 66.3 & 67.0 & 60.7 & 61.4 & 66.3 \\ 
  \rowcolor[HTML]{D9EAD3}
  \textbf{PRISM}      & \textbf{92.8} & \textbf{91.2} & \textbf{92.8} & \textbf{91.8} & \textbf{92.8} \\ 
  \bottomrule
\end{tabular}}
\vspace{-0.5em}
\caption{3-class classification results on NewsSpectrum.}
\label{tab:newsspectrum_3class}
\vspace{-0.5em}
\end{table}

\paragraph{Aligning Classification Labeling Scheme}
To facilitate consistent evaluation across datasets, we align the label schemes used in NewsSpectrum and BigNews. 
While NewsSpectrum employs a five-point ideological scale (\textit{Left}, \textit{Lean Left}, \textit{Center}, \textit{Lean Right}, \textit{Right}), BigNews uses a simpler three-class schema (\textit{Left}, \textit{Center}, \textit{Right}). 
We consolidate the five-point labels by mapping both \textit{Left} and \textit{Lean Left} to \textit{Left}, and similarly merging \textit{Right} and \textit{Lean Right} into \textit{Right}.
Table~\ref{tab:newsspectrum_3class} reports the performance under this aligned three-label setting, where PRISM continues to show strong results, confirming its stability under varying label granularities.

\begin{table}[t]
\centering
\small
\setlength\tabcolsep{4pt}
\renewcommand{\arraystretch}{1.5}
\resizebox{0.99\columnwidth}{!}{
\begin{tabular}{cccccc}
  \toprule
  \rowcolor[HTML]{FFF2CC}
  \textbf{Model} & \textbf{Acc $\uparrow$} & \textbf{Pre $\uparrow$} & \textbf{Rec $\uparrow$} & \textbf{F1-Ma $\uparrow$} & \textbf{F1-Mi $\uparrow$} \\
  \midrule
  \textbf{Llama-3.1-8B} & 27.6 & 37.8 & 26.9 & 21.6 & 27.6 \\ 
  \textbf{GPT-4o-mini}  & 32.7 & 45.4 & 32.5 & 30.1 & 32.7 \\ 
  \textbf{GPT-4o}       & 39.1 & 49.8 & 38.7 & 38.7 & 39.1 \\ 
  \rowcolor[HTML]{D9EAD3}
  \textbf{PRISM}        & \textbf{86.1} & \textbf{86.5} & \textbf{86.2} & \textbf{86.2} & \textbf{86.1} \\ 
  \bottomrule
\end{tabular}}
\vspace{-0.5em}
\caption{Classification results on NewsSpectrum with three LLM zero-shot learning baselines.}
\label{tab:LLM}
\end{table}

\paragraph{LLM Zero-shot Learning Baselines}
To contextualize PRISM's performance, we evaluate zero-shot political bias classification using powerful LLMs, including Llama-3.1-8B, GPT-4o-mini, and GPT-4o, on the NewsSpectrum dataset.

As shown in Table~\ref{tab:LLM}, while LLMs exhibit moderate performance out of the box, PRISM significantly outperforms them, particularly in Classification Accuracy (Acc) and F1-Macro (F1-Ma).
These findings underscore the value of PRISM's domain-specific architecture: although LLMs are versatile, PRISM yields more accurate and interpretable results through its targeted representation of political bias.

\end{document}